# The Indian Spontaneous Expression Database for Emotion Recognition

S L Happy, *Student Member, IEEE,* Priyadarshi Patnaik, Aurobinda Routray, *Member, IEEE,* and Rajlakshmi Guha

**Abstract**— Automatic recognition of spontaneous facial expressions is a major challenge in the field of affective computing. Head rotation, face pose, illumination variation, occlusion etc. are the attributes that increase the complexity of recognition of spontaneous expressions in practical applications. Effective recognition of expressions depends significantly on the quality of the database used. Most well-known facial expression databases consist of posed expressions. However, currently there is a huge demand for spontaneous expression databases for the pragmatic implementation of the facial expression recognition algorithms. In this paper, we propose and establish a new facial expression database containing spontaneous expressions of both male and female participants of Indian origin. The database consists of 428 segmented video clips of the spontaneous facial expressions of 50 participants. In our experiment, emotions were induced among the participants by using emotional videos and simultaneously their self-ratings were collected for each experienced emotion. Facial expression clips were annotated carefully by four trained decoders, which were further validated by the nature of stimuli used and self-report of emotions. An extensive analysis was carried out on the database using several machine learning algorithms and the results are provided for future reference. Such a spontaneous database will help in the development and validation of algorithms for recognition of spontaneous expressions.

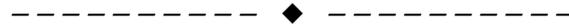

## 1 INTRODUCTION

Facial expression is a fundamental mode of communicating human emotions. Among the various channels that communicate human emotions – voice, textual content, gestures and facial expressions – it is considered one of the most accurate [1]. Hence, its relevance in the context of Human-Computer Interaction (HCI) is of primary importance. Automatic recognition of emotions through facial expressions, thus, becomes relevant for HCI as well as a number of other contexts [2] such as human monitoring, intelligent assistance, surveillance and so on. Tagging of affective states in the media content can be helpful in quick categorization of a large number of media files for automatic and effective content retrieval. Design of such systems [3] depend on reliable facial expression databases which include a number of parameters such as posed vs. natural expressions, gender, cultural variations, expression intensity variation, face pose, occlusion, contexts in which emotions are generated, etc. Well-labeled videos of facial behavior and emotion tagged images in the databases are, thus, of crucial importance. These are essential for training, testing and validation of algorithms for development of robust systems. Besides, recent research focuses on discrimination of fake expressions from the genuine ones [4]; all these provide a strong reason to create databases for high resolution videos of naturally elicited emotions.

Creation of an emotion database is a difficult and time consuming task [5]. However, database creation is an essential step in the creation of a system that will recognize human emotions. Spontaneous emotion elicitation requires significant effort in the selection of proper stimuli which can lead to rich display of intended emotions. Secondly, the process involves tagging of emotions by trained individuals manually which makes the databases highly reliable. Since perception of expressions and their intensity is subjective in nature, such expert ratings are essential for the purpose of validation.

Paul Ekman has reported the presence of six universal categories of facial expressions of emotions based on a number of cross-cultural studies [6] [7], regardless of cultural differences. Though a trained eye can recognise the facial expressions of people around the globe, it is very challenging for a machine to recognize the emotional states automatically. Machines recognize expressions based on the features extracted from the face such as shape, appearance etc. Due to variations in texture and shape of faces of people from different cultures and races, the advancement of affective computing relies on databases of faces from different ethnicity [8]. To the knowledge of the authors, no such emotion database exists for Indian faces. Moreover, the majority of the existing databases are created under tightly controlled laboratory environments [9] which do not represent the real world expressions that we come

- *S L Happy and A. Routray are with the Department of Electrical Engineering, Indian Institute of Technology, Kharagpur, India. E-mail: {happy,aroutray}@iitkgp.ac.in.*
- *P. Patnaik is with the Department of Humanities and Social Sciences, Indian Institute of Technology, Kharagpur, India. E-mail: bapi@hss.iitkgp.ernet.in.*
- *R. Guha is a psychologist in the counselling centre, Indian Institute of Technology, Kharagpur, India. E-mail: rajlakshmi_guha@rediffmail.com.*

across in our daily lives.

For reliable analysis of facial expression recognition algorithms, an Indian Spontaneous Expression Database (ISED) has been reported in this paper. The details of the experiment design for simultaneous elicitation and capturing of genuine expressions, its protocol, and annotation of emotion in the database are described in this paper. The performance of different expression recognition algorithms on this database are studied in detail. Strategies used for addressing the challenges for database creation are provided explicitly. In the database, emotions were induced through emotion-inducing videos in participants, along with simultaneous recording of videos at high frame rates. Basically, four kind of emotions are elicited and recorded, namely, (a) happiness, (b) sadness, (c) surprise, and (d) disgust. The emotional responses are segmented manually and annotated by trained decoders. Information such as expression intensity, gender, type of occlusion etc. are provided. Position of face, eye and nose in peak expression frames are also included in the database. This database, consisting of videos of full-blown as well as controlled facial expressions, will prove suitable for development of robust algorithms for automatic and non-intrusive detection of emotions from visual cues. To reduce intrusion and guaranty spontaneous emotion elicitation, concealed cameras were used without prior knowledge of the subjects and recording of physiological signals such as Electro-Encephalograph (EEG), Electro-Cardiograph (ECG) etc. were avoided.

The paper is organized as follows. Section 2 presents a review of earlier works. Section 3 presents the protocol followed during the experiments. Section 4 discusses the database content and its availability for research purpose. The evaluation of database is carried out in section 5. Section 6 concludes the paper.

## 2 BRIEF REVIEW OF EXISTING EXPRESSION DATABASES

Research on automatic affect recognition received an impetus around the late 1990s with early efforts of Mase [10] and Kobayashi [11] toward affect recognition from facial images. Since then, a number of emotion databases have been developed to advance the automatic recognition of affective states. In 1997, Lyons *et al.* [12] created the Japanese Female Facial Expressions (JAFFE) database which consists of 213 images of seven different emotional facial expressions (Sadness, happiness, surprise, anger, disgust, fear, and neutral) by 10 subjects. This database also includes the averaged semantic ratings of 60 Japanese female students on a scale of five. Another widely used database in this field is Cohn-Kanade AU-Coded Facial Expression database [13] which includes 486 sequences by 97 posers. Each sequence includes proceeding of a peak expression from a neutral expression. FACS coded peak expression and an emotion label for each sequence is provided in this database. However, these databases comprise posed expressions only.

Literature [14] emphasises the use of dynamic information in temporal domain — the changes in facial muscles with time — for accurately classifying emotions than the static data available in images. New efforts to create audiovisual databases include the work of [15], [16], [17], and [18]. Valstar and Pantic [19] have reported MMI database in which video clips with both frontal and profile views are included for posed expressions. The posed data include 20 participants showing 31 action units and 28 participants expressing six basic emotions. Annotation of the emotion and intensity has been included in the database for proper utilization of the database.

The necessity for spontaneous facial expression databases is emphasized in [8]. Excluding some attempts ([17], [20], [5], [21], [22], [23], [24]), most of the publicly available emotion databases include posed facial expressions only. In posed expression databases, the subjects are asked to display different basic emotional expressions, while in spontaneous expression database, the expressions are natural. The need for real time detection of emotion in practical contexts has encouraged researchers to create spontaneous emotion databases. The Extended Cohn-Kanade Dataset (CK+) [25] is an extension of [13] which includes posed expressions of 123 multi-ethnicity subjects along with the spontaneous smile expressions. The Geneva Multimodal Emotion Portrayal (GEMEP) corpus [26] includes the FACS coded emotion portrayals by trained actors which, though not spontaneous, are very close to the spontaneous expressions.

Works of Ekman [27] correlated the facial muscle movements with the underlying emotional state. Facial Action Coding System (FACS) measures all visually observable facial movements in terms of Action Units (AUs). Using this concept of facial muscle movements, posed expression databases include the reproduction of different emotion which are unnatural and hardly found in actual situations [17]. Spontaneous expressions differ from posed ones remarkably in terms of intensity, configuration, and duration. Apart from this, synthesis of some AUs are barely achievable [28] without undergoing the associated emotional state. Therefore, in most cases, the posed expressions are exaggerated, while the spontaneous ones are subtle and differ in appearance. In [29], high accuracy levels in automatic detection of posed expressions are reported; but accuracy for spontaneous expressions is still low. Because of the difficulty in elicitation of facial expressions and rigorous, time–consuming manual labour for the annotation of underlying emotional state in video clips, only a few spontaneous databases exist so far. Spontaneous expressions are very subtle, short lived, and often a mixture of pure expressions. So it is very difficult to reliably annotate the complex natural expressions. Recent efforts in HCI include understanding human emotions using multimodal signals such as voice, facial expression, posture, gesture, etc. Further, the relation between facial expressions and affective state is also necessary for proper assessment of cognitive states. The limited number of databases has always been an issue for researchers working in the field of emotion recognition. An ideal corpus should include high quality recording of rich emotional experience from different modalities along with the tagging of the emotional content.

Table 1
SUMMARY OF DATABASE CONTENT

| | |
|---|---|
| Number of video clips | 428 |
| Number of participants | 50 (29 male, 21 female) |
| Emotions elicited | Happiness (227 clips) Surprise (73 clips) Sadness (48 clips) Disgust (80 clips) |
| Clip duration | 1-10 sec |
| Clip selection | Manual |
| Self-report of emotion | Yes |
| Emotion rating scale | 0-5 (0: no emotion, 5:maximum intensity) |

In Belfast induced emotions database [17], emotions are induced using laboratory-based tasks and continuous trace ratings of the coloured responses are generated in terms of valence and intensity scale. It uses both active and passive tasks (watching emotional videos) to engage the subjects and thereby elicit emotion. In [15], to observe the range of emotional behaviour in a learning environment, non-verbal behaviour from cues such as facial expressions, eye-gaze and head posture data are collected to assess affective states during interaction and learning. In [16], spontaneous emotions are induced by activities such as playing computer games and conducting adaptive intelligence tests. Wang et al. [5] have developed a natural emotional database, containing both visible and infrared images, induced by film clips. They have recorded both spontaneous and posed expressions in three varied illumination conditions. In [18], the authors have used strategies to induce different emotions by exposing the subjects to different situations.

When creating the spontaneous facial expression databases, it is necessary to validate that the facial expression of a person is corresponding to the emotional state of the person. To create such an authentic validated database, Sebe et al. [30] have indicated some guidelines. The subjects should not be aware of being tested for elicitation of their emotional states; otherwise it influences their emotional states. Secondly, the subject's self-report should be documented after the test to validate the emotional states of the subjects. Thirdly, the presence of the experimenter may influence the elicitation of facial expression. Spontaneous emotion elicitation is possible through human to human interaction, human to computer interaction, by emotion eliciting tasks or by induction through picture, music, or videos [20].

However, in [15], [16], [17], [18] and [5], the subjects were aware of being monitored and the experiments were conducted in the presence of the experimenter, thus, violating the two guidelines mentioned by [30]. Hence the elicited emotion may not be purely spontaneous due to the possibility of social masking. In [31], the authors have created face database including facial expression videos elicited by watching emotional video clips. However, the annotation of the emotional state is not provided. VAM audio-visual database [22] is another spontaneous database where the clips are segmented form a German talk show and annotated using valence-activation scale. However, this does not permit self-report, nor the ability to elicit pure emotions. In [32], creation of a multimodal spontaneous emotion database is reported where audio-visual signals are captured along with feet pressure signal, thermal image, and body gesture by eliciting emotion by displaying pictures and videos as well as through interview. Recently, DEAP database [33] has been created which includes face videos of 22 participants along with physiological signal recording such as electro-encephalograph (EEG), electro-myograph (EMG), electro-oculograph (EOG), blood volume pulse (BVP), skin temperature, and Galvanic Skin Response (GSR). However, the more the number of signals recorded during the experiments, the more the chances of failure in eliciting natural expressions. Both the above databases suffer from excessive intrusion.

DISFA [21] includes full intensity FACS coding expressions of 27 participants elicited by viewing a video clip of duration of 4 minutes. Videos were captured at 20 fps at 1024x768 resolution. AU annotation for each frame is included in the database. However, the database has no description about elicitation of pure emotions. HUMAINE database [23] contains spontaneous video clips through different activities and conversations. It is a multimodal database including audio-visual data, physiological recordings (ECG, GSR, skin temperature, breathing, EMG and BVP) and performance data. Both global as well as frame-by-frame emotion label is provided. Conversations with individual SAL characters were used as the stimulus in the SEMAINE database [24] and the video was recorded at a spatial resolution of 780x580 pixels at 50 fps. The recordings of 150 participants are annotated on valence-arousal scale. In the above said databases, the participants were aware of being recorded. Moreover, the self-assessment were absent in the above said databases, so it is difficult to establish the ground truth.

The proposed experiment design takes care of all the above limitations and is in accordance with the guidelines developed by Sebe et al. [30]. The strategy of self-report of emotion and the use of hidden camera to record spontaneous expressions are adapted only in very few databases [30], [34]. Therefore, genuine expressions are captured during the experiments unlike the experiments in [17], [5], [24], [32], [33].

In practical scenarios, there is a possibility of observing mixed emotions. However, in this experiment, the passive mode of emotion elicitation provides scope for elicitation of pure emotions. The ISED contains high quality near frontal face recordings of spontaneous emotions at high resolution and frame rates along with information regarding gender of the participants, the ground-truth of emotional clips and its intensity, and the peak emotion intensity frame in the video clips. The head, the eyes and the nose positions in the peak intensity frames are also provided for easy access of the database. It covers the expressions of participants from different parts of India and from different cultural and linguistic backgrounds. Emotion elicitation was carried out without the presence of the ex-perimeter and recorded by a hidden camera; hence, spontaneous expressions were recorded. The facial expression annotation was conducted by four trained decoders of both gender. All these enhance the reliability of the database. Finally, the database is evaluated using baseline algorithms as an elementary assessment of usefulness of the database.

## 3 CREATION OF THE DATABASE

The ISED includes video and still facial images of spontaneous emotions. The emotion were elicited through passive elicitation by watching emotion inducing videos. Subjects were left alone in the experimental room and a hidden camera recorded their facial expressions. All the video clips included in the database vary from one to 10 seconds in length. The details of the database are provided in Table 1. The details of the videos used for emotion elicitation are provided in Table 2.

While constructing the ISED, the spontaneous emotion induction methods were administered carefully in specially designed artificial environments. Hence, the elicited emotions, though induced, look very natural and can be referred to as natural emotional expressions [35]. As expected in any experiment involving human subjects, the effectiveness of the used stimuli in emotion induction can have wide variance based on subjective perception and the state of mind of the subjects. For example, a specific video clip was reported as very disgusting by most of the subjects, while a few reported it to be fearful or sad. Similarly, the intensity of emotion elicitation varied for individuals for the same video. However, as observed from Table 2, the video clips were successful in eliciting the targeted emotion in most of the participants. In cases where the self-report, type of stimuli and decoders' assessments showed significant variances, the clips were not included in the ISED.

The facial expressions of emotions present in the recorded videos were extracted by specially trained decoders selected by screening more than a 100 prospective decoders. Four decoders, two of each gender, were selected on the basis of their ability to detect emotions on 28 multi-ethnic images of faces (Japanese and Caucasian Facial Expressions of Emotion database [36] images) exposed for 1/15th of a second. Their success rates were found to be above 95%. They were later trained using Micro Expression Training Tool and Subtle Expression Training Tool [37] until they achieved an accuracy rate of more than 90%. They were also familiarized with Facial Action Coding System.

### 3.1 Experiment Setup

Forced or posed expressions can be captured by voluntary participation of the subjects. However, as discussed earlier, for spontaneous emotion elicitation and capture of such data, awareness of the experiments' intents and presence can be a barrier in elicitation of spontaneous emotions [30]. The subjects' awareness about experimental intent can be detrimental. Secondly, video documentation done without the knowledge of the subjects would again aid spontaneous emotion manifestation. Awareness of the presence of camera is similar to the presence of an experimenter and subjects often mask their emotions. Thirdly, such experiments should ideally be conducted indoors where noise, movement, and other such distractions can be controlled and proper ambient lighting maintained for quality recording of videos. Based on the suggestions of subjects during pilot studies we concluded that directly lighting the subjects made them suspicious and conscious and also disturbed their viewing experience. Hence, ambient reflected lighting was used.

The whole experiment was set up in an isolated room of area 3mx3m with a single door. As discussed earlier, for elicitation of natural emotions, the participants were not informed about the goal of the experiments beforehand and the video were captured using a hidden camera. To conceal the camera, a wooden box was built with a window made of one-way glasses so that the camera could capture videos of the subject if placed inside the box, but the subject would not be aware of its presence. Two reflected light sources were used to create ambient lighting conditions, thereby providing a good environment to watch movies from participants' point of view and enough lighting to capture good quality videos from experiment point of view. The camera was concealed in the wooden box as shown in the Fig. 1.

During passive emotion elicitation, a monitor was put on a table in front of the concealed camera box. Participants were asked to sit comfortably on a height-adjustable movable chair and watch the video/film clips. In order to provide privacy to the encoders for full blown emotion induction, the subjects were allowed to watch videos alone in the experimental room. To avoid suspicion subjects were allowed to sit at their own ease without any other restrictions inside the experimental room. The doors and windows were kept shut during the entire period of the experiment to avoid external interference.

### 3.2 Camera Setup

A Nikon D-5200 camera was used for video recording which can record videos at 50 frames per second with a resolution of 1920x1080. A Nikon D-7000 was used as a backup which could also record with the same resolution.

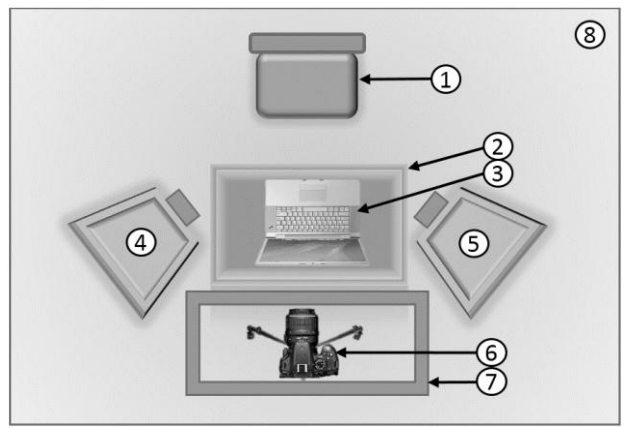

Fig. 1. The Experimental Set-up. (1) chair for subject, (2) table, (3) laptop, (4-5) light sources, (6) camera on tripod, (7) the wooden box to conceal camera, (8) experiment room

TABLE 2
DETAILS OF EMOTIONAL VIDEO CLIPS USED FOR EMOTION ELICITATION

| Content of Video clips | Percentage of participants labeled the video clips into different categories in self-assessment report (values in %) | | | | | | Duration |
|---|---|---|---|---|---|---|---|
| | Happiness | Surprise | Sadness | Disgust | Fear | Anger | |
| Robin (EEF) | 86 | 14 | 0 | 16 | 0 | 0 | 2:32 |
| A scene from Jaane Bhi Do Yaaron [Bollywood movie - 1983] | 88 | 14 | 14 | 0 | 0 | 0 | 3:13 |
| Disney Earth (EMDB) | - | - | - | - | - | - | 1:03 |
| Unbelievable stunts | 4 | 96 | 0 | 0 | 12 | 0 | 1:43 |
| Scary accident compilations | 0 | 70 | 10 | 0 | 30 | 2 | 2:04 |
| Diary of a Nymphomaniac (EMDB) | 0 | 2 | 82 | 0 | 24 | 0 | 1:48 |
| Very sad movie clip | 0 | 0 | 100 | 0 | 2 | 6 | 4:03 |
| Disney Earth (EMDB) | - | - | - | - | - | - | 0:31 |
| Pink Flamingos (EEF) | 2 | 0 | 4 | 94 | 0 | 0 | 0:37 |
| Eating live cockroaches | 0 | 2 | 4 | 94 | 6 | 4 | 1:53 |

*The videos used from the reported literatures are mentioned in the brackets. Here EMDB stands for "The Emotional Movie Database" [38] and EEF stands for "Emotion Elicitation Using Films" [39]. Note that the participants were allowed to assign multiple emotional labels to one video clip.*

The camera was placed inside the wooden box concealed from the subjects at a height of 1.5 m from the ground. The distance between the subject and the camera was approximately 0.8m to 1.5m.

### 3.3 Illumination Setup

Proper lighting is essential while recording videos at high frame. Due to very short image acquisition time, unless the light intensity is considerably high, the captured videos looks dark. To address this issue, a pair of studio lights were bounced off the walls in the experiment room to create suitable ambient illumination and in order to reduce the discomfort of directing light into the face of subjects. The subjects were informed that the soft ambient light was used to create a pleasant and comfortable viewing atmosphere.

### 3.4 Subjects

Fifty healthy participants, 29 males and 21 females, voluntarily participated in the experiment. The participants were of the age group ranging from 18 to 22 years and were from different regions of India. We administered General Health Questionnaire (GHQ) and State-Trait Anxiety Inventory (STAI) prior to the experiments for screening for physical illness or mental distress. They were debriefed at the end of the experiments. The database includes the videos of the participants who signed the consent form agreeing to their facial videos being used for research purpose. The participants were from different states, different backgrounds, and different cultures. Of them, 38% belong to north zone of India, 24% are from east zone, 22% are from south zone and 16% are from west zone of India.

### 3.5 Stimuli

Video clips (18 seconds to 4.5 minutes in duration) identified on the basis of pilot studies (as well as clips used in earlier studies such as The Emotional Movie Database [38] and Emotion Elicitation Using Films [39] etc.) were used to elicit emotions. Four emotions were elicited, namely, happiness, surprise, sadness, and disgust.

Carvalho *et al.* [38] reported that the video clips are more effective stimuli than the other stimuli for evoking emotional responses. Selection of emotional videos is an important step as it affects the strength of emotion that one can feel. The emotional videos were collected from different databases that have been used for similar studies (viz. [38], [39], [40], [41]) and some of the selected videos were from other sources such as Bollywood films and YouTube.

Since the participants were all Indian, the videos were selected carefully to elicit emotions for the target demography. Ten individuals including the decoders were asked to watch and rate the collected videos. The rating was carried out on a scale of 0-5 for different emotions where 0 indicates no emotion at all and 5 indicates very high intensity of elicited emotion. The average rating for each video was calculated. Some videos were found to elicit multiple emotions. Videos eliciting multiple emotions may lead to a mixed emotional expression, and hence were rejected. The videos which were unanimous in eliciting one specific emotion with higher intensity were selected. Keeping in mind the Indian cultural context, for each of the four emotions we used one video clip suggested by the databases and the other from non-database videos, and in each case selected the one that had got the highest ratings from our decoders. Including two neutral video clips, 10 such video clips were selected for different emotions such as happiness, surprise, sadness, and disgust. The agreement between the coders was found to be 0.81 by computing Fleiss's kappa [42]. Fleiss's kappa of more than 0.8 is considered as very reliable degree of agreement. Since eliciting anger fear is difficult in passive elicitation, it was not considered. Similarly, fear elicitation was also not carried out because of ethical issues. The length of the selected clips varies from 18 seconds to 4.5 minutes which is in accordance with the recommended length of emotional video as described in [39]. The happiness and surprise videos were shorter, while the sadness inducing videos were longer as it takes more time to build up sadness. The selected videos were strong enough to elicit moderate emotions, but not cause ethical concern.

### 3.6 Self-Report of Emotion

Lack of accurate assessment of emotional experience in the subjects can jeopardize an experiment. To identify the emotion experienced by the subjects, one may directly ask them or infer from physiological signals. Since our design was non-intrusive, self-report forms were included for subjective assessment of emotions. Instead of telling anything about the experiment, the subjects were informed that they had to participate in a survey where they had to assess the emotional content and intensity of the video clips; and for accurate assessment of emotions they had to get immersed in the viewing experience and experience the emotions. At the end of each video clip, they were informed to note down the emotions that the video clip communicated and the intensity with which they felt them on a scale of zero to five. They were given a list of six basic emotions [43], but were also told that they could add other emotions if they wanted. Hence, the self-report of emotion was gathered without the subjects knowing that they were being assessed. The average emotion reported by the participants in their self-assessment report gives an insight to the expressivity of the subjects to different emotion categories which is provided in Fig. 2. Compared to the other emotions, happiness videos were rated slightly lower. On the other hand, the disgust videos were very successful at eliciting the targeted emotion.

### 3.7 Occlusions

Presence of an obstacle such as spectacles, hair, beard, moustache etc. disguises some of the important information in facial expression which is often a challenge for practical implementation. Therefore, for the experiments, we decided to include some subjects who wore spectacles. Some of the male subjects also had moustaches and beards. They were also not restricted from touching their chins or cheeks since this also lead to partial obstruction which we intend to capture.

### 3.8 Experiment Procedure

The volunteers were asked to fill up GHQ and STAI and screened before being allowed to participate in the experiments. The participants were made aware of their rights to withdraw from the experiment at any time they desire. As mentioned earlier, their task was to watch the video clips, identify the emotions expressed and rate the intensity of emotions. Therefore, they were taken to the experiment room and left alone with a rating sheet where they could evaluate the emotions they experienced after watching each video. The camera was switched on before the subject entered the experiment room. During debriefing it was found that none of the subjects had detected the concealed camera or suspected any other mode of recording. We observed some persons make the recordings of face difficult by occluding their face, leaning forward, or looking away from screen (mostly during disgust) when watching the videos. Hence, the arrangement of table, arm-rest and distance of viewing were altered to control this without making the subject suspicious by explicit instructions. However, they were encouraged not to close their eyes during disgust or other such negative emotions so that their ratings would be accurate. The self-report of emotion was generated as described earlier. A gap of 15 seconds was provided between subsequent clips for rating of the previously shown clip.

The videos were shown in two sessions, of 12-14 minutes duration each. The sequence in which the emotion induction clips were presented was happiness, surprise, sadness and disgust. The first session included happiness, and surprise followed by a break of five minutes. The second session includes videos of sadness and disgust. In the beginning, we showed a funny video to make the participant comfortable with the experiment environment. Another amusement video was used at the end of the second session in order to dispel the feelings of sadness and disgust and make the participants happy. To avoid the sudden transition of one emotional clip to another, some neutral videos were introduced between two different emotion clips. Throughout the experiment, the order of the videos was kept the same.

After the experiments, the subjects were taken to a different room where they were debriefed. They were made aware of the project and the requirement of spontaneous expression database for improvement of computer vision algorithms in emotion recognition and requested to provide a written consent for use of the recordings for academic and research purpose. Their questions regarding the database, the video content, and the projects were clarified. They were not pressurized in any way to consent to the use of their in the database. A few participants refused to contribute to the database. In all such cases their video clips were permanently destroyed.

### 3.9 Ethics statement

The experimental procedure and the video content shown to the participants were approved by the Institutional Ethics Committee (IEC) of IIT Kharagpur. The participants were also informed that they had the right to quit the experiment at any time. The video recordings of the subjects were included in the database only after they gave a written consent for the use of their videos for research

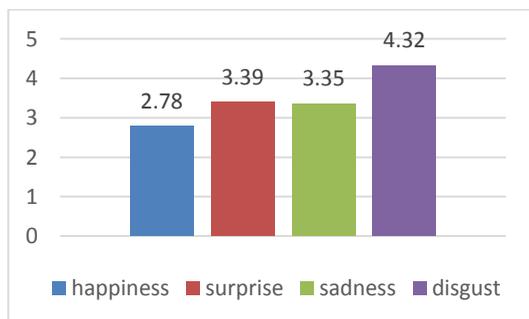

Fig. 2. The average emotion intensity reported by the participants to different emotion categories

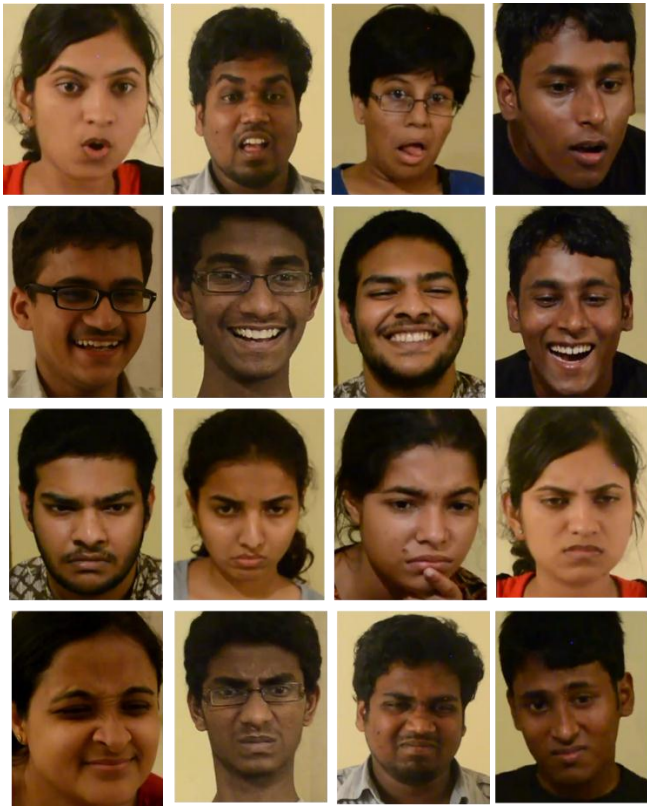

Fig. 3. Peak intensity images of different spontaneous expressions, surprise (1st row), happiness (2nd row), sadness (3rd row) and disgust (4th row)

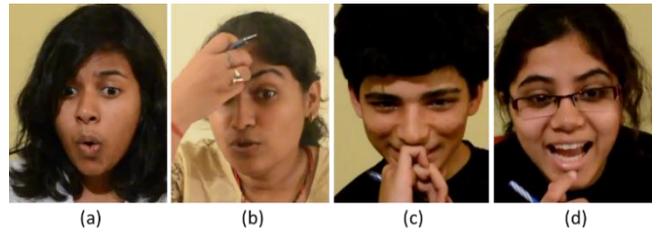

Fig. 4. Example images of occlusion

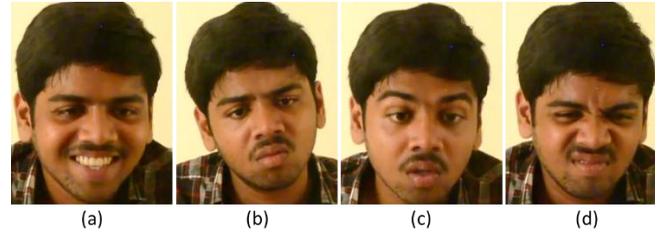

Fig. 5. Different expressions of a single participant

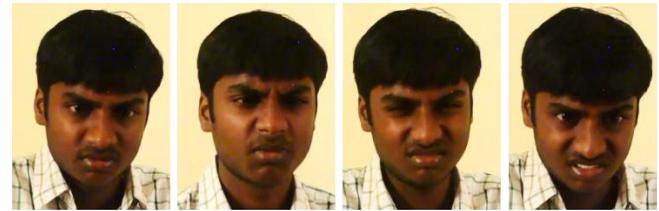

Fig. 6. Variation of intensity of the peak images in different video clips of same type of expression (disgust) of same participant

purpose. A few subjects were also agreed to use their face images in research articles.

## 4   DATABASE CONTENT

The created database contains subtle to full blown elicitation of different emotions. It contains facial expression videos and still images of 50 participants with the emotion ground truth and its intensity. A few examples of the database images are provided in Fig. 3. Different occlusions are shown in Fig. 4 and all expressions of the same participant are shown in Fig. 5.

The recording data for each participants was around 30 minutes long. The video clips were manually screened for facial expressions and segmented out. The length of the extracted clips vary from 1 sec to 10 sec. The average duration of the segmented clips are 4 seconds. We observed that the expression duration varied with the kind of emotion. For example, it is very difficult to get a small clip for sadness, because of its gradual evolution and persistence over a relatively long time. On the other hand, surprise expressions set very fast and disappeared in a moment. For expressions that lasted for a long time, we segmented out a short clip out of it labeling one emotion to it at the center of the clip. It was observed that most of the expressions were accompanied by head and body movements. All subjects did not show all types of expressions. However, we found happiness and disgust to be induced easily, while sadness was very difficult to induce. However, some participants showed spontaneous surprise easily through passive induction.

The segmented clips were annotated by four trained decoders to tag the six basic emotional expressions viz. happiness, sadness, surprise, fear, anger and disgust, and the corresponding intensity on a six-point scale. In the scale of 0-5, higher value in the scale corresponds to high intensity of elicited emotion. The clips which are classified to the same emotion by all decoders were considered for inclusion in basic emotion category. The intensity of such clips were decided by taking average of ratings of all decoders [5]. The reliability of agreement between four raters is also evaluated by using Fleiss's kappa [42]. Here the evaluation is based on the classification of the video clips into the categories of expressions. The Fleiss's kappa coefficient of the labeling was 0.847, indicating a very good consistency. Since the spontaneous expression of an individual does not remain the same with different situation and even varies across time, several expressions of the same participants are included in the database. Moreover, the face pose and the intensity of expression vary in different video clips of a participant while displaying the same expression. Fig. 6 shows the variation of expression intensity and face pose in different video clips. The emotion category was validated by the type of video the participants had watched during invoking that expression and the self-report of emotions. Thus, the segmented expression clips were included in the database based on agreement of type of stimuli used, ratings of the coders, and the self-report of emotion by the participants. Thus, we strongly believe that the videos included in our database are genuine and spontaneous. In some cases, we observed mixed emotion which was difficult to annotate to the basic emotion categories. In such cases, the clips were excluded from the database.

Most of the video clips start with a neutral face and end either with the peak expression or after the off-set of the expression. Thus, it provides the researchers a chance to model the transition of spontaneous facial expressions from the neutral face. However, the sadness clips were segmented around the peak expression frames. Thus, the neutral face may not be at the starting frame of a sadness clip.

### 4.1 Availability

The ISED database can be obtained by writing a mail at iseddatabase@gmail.com. The description of the database is available at https://sites.google.com/site/iseddatabase/. An End User License Agreement (EULA) needs to be produced for accessing the database.

## 5 EVALUATION OF THE DATABASE

This section describes the procedures that has been carried out to obtain the baseline results using the database. This elementary assessment provides an insight to the usability of the database. We conducted the expression recognition experiment using combination of different types of facial features and classifiers. The peak expression faces in all the video clips were used in the experiment. Though most of the images are near frontal, some are with face occlusion and head rotations. Among the selected peak images, 227, 73, 48, and 80 images belong to the class happiness, surprise, sadness, and disgust respectively.

Face detection is the first step in expression recognition. Viola-Jones Haar cascade classifiers were used for face detection which resulted an accuracy of 89.72% in our dataset. The presence of off-plane head rotation and occlusion are the primary reasons behind the poor performance of the face detector. However, for establishing the baseline, the face position was selected manually in the images where the face detector failed. Similar techniques were adopted for eye and nose localization. The eyes and nose positions were also manually marked in case the Haar classifier failed. The ground truth of face position, eye positions and nose position in all peak images are also included in the database.

The eye locations were used to suppress the effect of scale, position and in-plane rotation. The face image was first rotated to bring the eyes to the same horizontal level. Then the image was scaled so that the eye centers were positioned at a particular distance. The face region was extracted in all the images and resized to a standard resolution of 96x96. The face images were converted into grayscale images for further processing. A Gaussian mask was applied to the whole image to remove noise followed by histogram equalization. Ten-fold cross validation was adopted in our experiments.

### 5.1 Feature Extraction

The accuracy of expression recognition highly depends upon the selection of appropriate features to represent the expressive face. The geometric or the appearance features or the combination of both can be used for this purpose. We have experimented with different types of feature extraction techniques as described below.

#### 5.1.1 Grayscale Intensities

After preprocessing of the face image, the pixel intensities of the gray image were used as feature vectors for classifying expressions [5]. Since the pixel intensities vary from 0~255, they were normalized to zero mean and unit variance.

In another set of experiments, the eye region and the lips region was segmented and the corresponding pixel intensities were used for expression classification [44]. The eye region was selected based on the positions of the eyes. Similarly, the lips region was determined with respect to the position of the nose. The size of the selected regions depend upon the size of the face. In Fig. 7, the process for selection of eye and mouth regions is provided.

#### 5.1.2 Local Binary Pattern (LBP)

LBP operator compares the pixel values in a location to its neighboring pixel values and generates a binary number representing the pattern [45]. The histograms of LBP image forms a robust feature descriptor against illumination variation. We used LBP histograms ($LBP_{P,R}$), uniform LBP ($LBP_{P,R}^{u2}$) histograms, and rotational invariant uniform LBP patterns ($LBP_{P,R}^{riu2}$) in our experiments.

The LBP histogram represents the patterns of the whole image which lacks the spatial information. The LBP histograms from the image sub regions can be concatenated to capture the details of the position of the patterns. In our experiments, we divided the face image into 3x3, 5x5 and 7x6 non-overlapping regions to construct the enhanced feature vector. Fig. 8 describes the procedure of multi-block LBP feature extraction technique where local features are extracted by dividing the face into $m \times n$ sub regions.

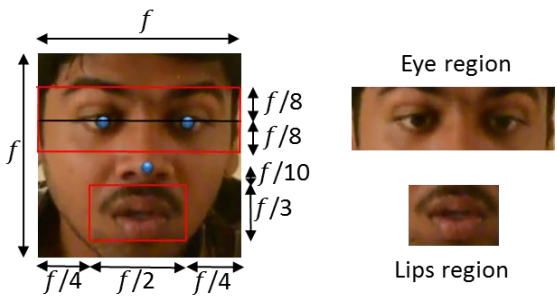

Fig. 7. Selection of lips and eye regions

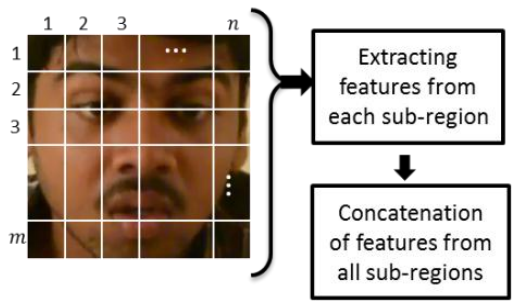

Fig. 8. Multi-block feature extraction using $m \times n$ number of regions

Table 3
FACIAL EXPRESSION RECOGNITION RESULTS OBTAINED USING
DIFFERENT METHODS WITH TEN-FOLD CROSS VALIDATION

| Feature | Method | Precision | Recall | F1 score | Average accuracy (in %) |
|---|---|---|---|---|---|
| Grey intensities of the whole face | PCA + KNN | 0.5599 | 0.5365 | 0.5480 | 64.48 |
| | PCA + LDA | 0.6865 | 0.7015 | 0.6939 | 75.70 |
| | PCA + Naïve-Bayes | 0.6429 | 0.6097 | 0.6259 | 69.62 |
| Grey intensities of lips and eye region | PCA + KNN | 0.6812 | 0.5995 | 0.6378 | 71.02 |
| | PCA + LDA | 0.6330 | 0.6567 | 0.6447 | 69.39 |
| | PCA + Naïve-Bayes | 0.6454 | 0.5959 | 0.6197 | 69.85 |
| LBP features | $LBP_{8,1}$ from 7 × 6 Regions + PCA+LDA | 0.7942 | 0.7735 | 0.7837 | 82.47 |
| | $LBP_{8,1}$ from 5 × 5 Regions + PCA+LDA | 0.7648 | 0.7737 | 0.7692 | 81.54 |
| | $LBP_{8,1}^{u2}$ from 7 × 6 Regions + PCA+LDA | 0.7078 | 0.7150 | 0.7114 | 77.80 |
| | $LBP_{8,1}^{u2}$ from 5 × 5 Regions + PCA+LDA | 0.7241 | 0.7335 | 0.7287 | 78.73 |
| Gabor filters (5 scale and 8 orientation) | PCA + LDA | 0.7881 | 0.7763 | 0.7822 | 82.00 |
| | Adaboost | 0.6429 | 0.5372 | 0.5853 | 69.62 |
| | PCA + KNN (Eye and lips regions only) | 0.6551 | 0.5826 | 0.6168 | 69.15 |
| | PCA + LDA (Eye and lips regions only) | 0.7477 | 0.7292 | 0.7383 | 78.03 |
| PHOG | PCA + LDA | 0.6226 | 0.6581 | 0.6398 | 65.88 |
| | Adaboost | 0.6082 | 0.3748 | 0.4638 | 60.74 |
| | PCA + LDA (Eye and lips regions only) | 0.6345 | 0.6708 | 0.6521 | 67.55 |
| | Adaboost (Eye and lips regions only) | 0.5468 | 0.3432 | 0.4217 | 57.47 |
| LGBP | With $LBP_{8,1}^{u2}$ from 3 × 3 Regions + PCA + LDA | 0.8510 | 0.7974 | **0.8233** | 86.44 |
| | With $LBP_{8,1}^{u2}$ from 5 × 5 Regions + PCA + LDA | 0.8332 | 0.7992 | 0.8158 | **86.46** |
| | With $LBP_{8,1}^{u2}$ from 5 × 5 Regions + PCA + LDA (Eye and lips regions) | 0.7660 | 0.6969 | 0.7298 | 78.50 |

### 5.1.3 Gabor Wavelets

Gabor wavelets [46] are well known for their ability to capture discriminative features with orientation specific frequency bands. The magnitude response obtained by Gabor filters of different orientation and spatial frequency can represent the image features in a single vector. In our implementation, we used five scales and eight orientations; thus forty Gabor filter banks in total.

### 5.1.4 Local Gabor Binary Pattern (LGBP)

The LGBP of an image is obtained by applying LBP operator on the Gabor magnitude images of different orientation and spatial frequency. Thus the multi-resolution and multi-orientation relation between the spatial frequency values are incorporated in the feature vector. A detailed study of facial expression recognition using LGBP features is provided in [47] and [48]. It has been proved that LGBP is very robust to illumination variation and misalignments.

In our experiments, LGBP features were extracted from 40 Gabor magnitude maps (5 scales and 8 orientations) by diving each Gabor maps into 3 × 3, 5 × 5 and 7 × 6 subregions. The uniform binary patterns ($LBP_{P,R}^{u2}$) and the rotation invariant uniform binary patterns ($LBP_{P,R}^{riu2}$) were used to encode the texture of the region.

### 5.1.5 PHOG descriptor

Object boundaries play vital role in object representation and identification. Histogram of oriented gradients (HOG) can represent the object geometry which is obtained by dividing the image into several blocks, calculating the histograms of edge directions in each block and concatenating the histograms of different blocks into a single shape descriptor. The angular range 0 ~ 180° can be quantized into several binwidths for obtaining the histogram of the edge orientations. Bosch et al. [49] proposed pyramid of histogram of gradients (PHOG) descriptors which represents the position of the orientations along with the HOG features. The spatial location of the object is encoded by calculating the HOG features of the object at different resolutions and concatenating them to form the feature vector. Thus PHOG features of an object are robust against the slight shape change, rotation and illumination variation in the image.

In our experiments, nine bin histograms were used for the angular orientations for each 8 × 8 cell to represent the object. A three layer pyramid structure was used for accurately representing the shape.

## 5.2 Expression Classification

Different classification techniques were adopted in our experiment following different literature. The eye and lips regions were extracted from the aligned face images as in [44]. The feature extraction techniques described in section 5.1 such as LBP, Gabor filters, LGBP, PHOG etc. were carried out on the whole face or on the eye and lips regions. The feature normalization was carried out to map the features to zero mean with unit variance.

Feature selection is an important step before feeding the data to a classifier. Principal Component Analysis (PCA) is widely used as a dimensionality reduction tool which reduces the dimension of the feature vector with minimal loss of information. Adaboost selects a fewer dimensions from the feature vector that provides significant accuracy during classification into different classes. Adaboost is also a fast classifier.

In our experiments, PCA was primarily used for reducing dimensionality of the feature vector. However, while applying Adaboost, PCA was not applied beforehand since Adaboost automatically selects a few features. In all scenarios with application of PCA, the number of principal components were selected in such a way that the original signal can be reconstructed with an error less than 5%.

We implemented PCA and PCA+LDA (Linear Discriminant Analysis) framework for grey intensity features as reported in [5]. In [50] and [51], the face region is divided into several sub-regions and the feature vector is constructed by concatenating the local features extracted from each sub region. In our experiments, we adopted this method to extract LBP related features. As described in [52], multiclass Adaboost was used for expression classification using features of whole face or parts of face. One-against-one multiclass Support Vector Machines (SVM) [53] with linear as well as radial basis function kernels were also used for classification. However, it is not reported due to its poor performance on our database. Cohen *et al.* [54] used the Naïve Bayes classifiers for emotion classification as it is reported to be successful in many practical problems. We have also implemented Naive Bayes and K Nearest Neighbor (KNN) classifiers in our experiments.

### 5.3 Results and Discussion

A series of experiments with different feature extraction techniques, dimensionality reduction techniques and various classifiers were conducted for analysis of ISED expression images. However, performances of a few selected combinations are provided in Table 3 which produced significant accuracy. The precision, recall and F1 score for each experimental method are also provided.

As observed from Table 3, the grey intensity of the face region achieved an average recognition rate of 75.5% by using PCA + LDA classifier. However, the performance of grey intensities of the eye and lips regions was relatively lower compared to the performance of whole face image indicating the fact that some of the important information were missing. This was also supported by the results of Gabor features. By applying Gabor filter banks with 5 scales and 8 orientations on the whole face image, the accuracy was observed to be 82% with an F1 score of 0.78. In contrast, the same Gabor features could achieve an accuracy of 78% while applied on lips and eye regions.

The LBP operators were found to be successful in expression classification for the ISED images. We experimented with various LBP features extracted from different number of regions of face image as shown in Table 4. The $LBP_{8,1}^{u2}$ features performed poorly in comparison to $LBP_{8,1}^{u2}$ and $LBP_{8,1}$ features. The $LBP_{8,1}$ features achieved the recognition rate of 82.47% which was the best among all the LBP based techniques. It is also clear from Table 4 that extraction of features from $7 \times 6$ number of regions improves the classification accuracy since the local features are encoded properly.

The PHOG features did not perform well when extracted from either the whole face or the specific facial regions. The reason behind the failure of the shape features may be explained by the fact that the database images are having different head rotations along with different intensity of expressions. Further, occlusions change the shape of the region. Thus, a general rule for associating certain shape features to an expression is difficult.

LGBP features outperformed the rest feature extraction techniques. We have implemented 5 scale and 8 orientation Gabor filter bank followed by $LBP_{8,1}^{u2}$ and $LBP_{8,1}^{riu2}$ for extracting LGBP features. As observed from Table 3, LGBP features encoded by $LBP_{8,1}^{u2}$ from $5 \times 5$ facial regions achived best recognition accuracy of 86.46%. On the other hand, based on F1 score, uniform pattern LGBP from $3 \times 3$ regions performed best with an F1 score of 0.8233. Table 5 displays the results obtained by dividing the Gabor maps

#### TABLE 4
#### The Performance of Different LBP Features Using PCA+LDA Classifier

| | Number of regions | | |
|---|---|---|---|
| | $3 \times 3$ | $5 \times 5$ | $7 \times 6$ |
| $LBP_{8,1}$ | 75.23 | 81.54 | **82.47** |
| $LBP_{8,1}^{u2}$ | 77.57 | 78.73 | 77.80 |
| $LBP_{8,1}^{riu2}$ | 60.28 | 62.14 | 71.72 |

#### TABLE 5
#### The Performance of Different LGBP Features Using PCA+LDA Classifier

| | Number of regions | | |
|---|---|---|---|
| | $3 \times 3$ | $5 \times 5$ | $7 \times 6$ |
| LGBP with $LBP_{8,1}^{u2}$ | 86.44 | **86.46** | 86.21 |
| LGBP with $LBP_{8,1}^{riu2}$ | 81.77 | 82.47 | 85.04 |

#### TABLE 6
#### The Confusion Matrix using PCA+LDA Using LGBP Features with $LBP_{8,1}^{u2}$ from $5 \times 5$ Number of Regions

| | Happiness | Surprise | Sadness | Disgust |
|---|---|---|---|---|
| **Happiness** | **0.9692** | 0.0044 | 0.0044 | 0.0220 |
| **Surprise** | 0.1096 | **0.7945** | 0.0822 | 0.0137 |
| **Sadness** | 0.0417 | 0.1042 | **0.7083** | 0.1458 |
| **Disgust** | 0.2125 | 0.0250 | 0.0375 | **0.7250** |

into different number of regions. The confusion matrix of the results obtained by using LGBP features with $LBP_{8,1}^{u2}$ from $5 \times 5$ number of regions is provided in Table 6. It was observed that the LGBP features with uniform patterns performed better than rotation invariant patterns. Further, the recognition rate of $LBP_{8,1}^{u2}$ was similar when extracted from different number of facial sub-regions. Therefore, minimum number of divisions are preferred to reduce the feature dimension.

Among the classifiers, LDA technique excelled in our experiments. Other classifiers such as multi class SVM, Adaboost, KNN, and Naive Bayes did not perform well compared to the performance of LDA technique. The poor performance of SVM in our database was probably because of the imbalanced dataset. Moreover, the number of training samples are less. Thus, LDA is the suitable tool for finding the hyper-plane that minimizes the intra-class scatter, while maximizing the inter-class scatter.

The best results obtained by both LBP and Gabor features are almost comparable. However, LGBP features yielded best results with an accuracy about 86% which is better than both LBP and Gabor. This may be explained by considering the fact that LGBP features combine the attributes of both LBP and Gabor. However, the computational cost of LGBP is very high compared to the individual features. In real-time applications, the LBP features may be used since its computational complexity is very low compared to other feature extraction techniques.

The best expression recognition performances of our baseline algorithms are still low which may further be improved by extracting features from specific regions of the face. Addressing the illumination variations and face occlusions may also be considered to improve the accuracy. The presence of wide variations in facial expressions and their intensities along with the occlusion, facial poses and arbitrary head movements are the practical scenarios which should be tackled for improving the facial expression recognition systems.

## 6 CONCLUSION

Emotion recognition using facial expressions - a medium for a natural way of communication with machines - needs spontaneous facial expression databases with reliable annotation of emotions. The ISED contains emotional responses in the Indian context and fulfils a number of aspects of the desired requirements. The database includes mild to strong spontaneous facial expressions. Its realistic nature can help researchers to develop algorithms for recognition of human emotions in practical situations. Several strategies are adopted for the creation of the database which have been briefly described. The tasks are designed to keep the subjects engaged and to induce spontaneous emotions. Since the expressions of an individual varies across time, several expressions of the same participants are included in the ISED. The data collected after recording of the experiments are further reduced to small video clips containing only the emotional expressions. The video clips of the database are annotated carefully by trained decoders, which are further validated by the self-report of emotion by the participants and the type of stimuli used. We observed significant elicitation of sadness and surprise in a few subjects, while most of the subjects displayed happiness and disgust expressions easily.

Some evaluation protocols were carried out to provide reference evaluation results for researchers for further improvement of the expression recognition techniques. The presence of unrestricted head movements and various face poses are the primary issues which may be addressed to improve the accuracy. Further, the face occlusions through spectacles, facial hair, hand etc. are also present in ISED which increase the complexity of expression recognition. The spontaneous facial expressions of ISED with a reliable ground truth would help the researchers to develop and validate their algorithms for practical applications.

## 7 ACKNOWLEDGMENT

The authors would like to thank all the participants for their participation in the creation of this database. The authors would also like to thank the team of volunteers for their helpful contribution in developing or collecting video materials used in the experiments. Special thanks go to the decoders who have helped us in manually annotating the database.

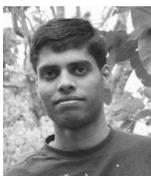

**S L Happy** has received the B.Tech. (Hons.) degree from Institute of Technical Education and Research (ITER), India in 2011. Now he is pursuing the joint MS – PhD degree at Indian Institute of Technology Kharagpur, India. His research interests include pattern recognition, computer vision and facial expression analysis.

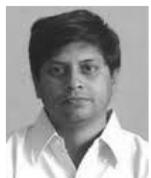

**Priyadarshi Patnaik** completed his Masters in English Literature in 1992 and his PhD in Indian Aesthetics (Indian Aesthetic Emotions) in 1995. He joined IIT Kharagpur in the year May 1997. His areas of research include Indian aesthetics (aesthetic emotions), visual culture and communication, cultural translation theory and practice, media and multimedia studies, and nonverbal communication. He is currently working as Professor at the Department of Humanities and Social Sciences, IIT Kharagpur. He has published about 30 research articles in national, international journals and edited books, written eight volumes of critical and creative writing on areas such as aesthetic emotions, Indian aesthetics, communication, translation and poetry, edited three books on aesthetics and aging, and has published more than 30 of his poems, visual arts, illustrations and photographs in international journals. In the past he has, in association with Defense Institute of Psychological Research, India, undertaken two projects on lie-detection using non-verbal cues and negotiation and interrogation using non-verbal cues.

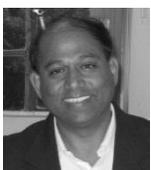

**Aurobinda Routray** has received his Masters degrees in 1991 from IIT Kanpur, India and his PhD in 1999 from Sambalpur University, India. He has also worked as a postdoctoral researcher at Purdue University, USA, during 2003-2004. He is currently working as a professor in the Department of Electrical Engineering, Indian Institute of Technology, Kharagpur. His research interests include non-linear and statistical signal processing, signal based fault detection and diagnosis, real time and embedded signal processing, numerical linear algebra, and data driven diagnostics.

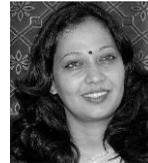

**Rajlakshmi Guha** is a psychologist in the counseling center, Indian Institute of Technology, Kharagpur, India since 2009. Before joining in IIT Kharagpur, she taught in the University of Calcutta and the West Bengal State University. She received her doctorate degree in Social Clinical Psychology from University of Calcutta. She has publications in the field of depression, stress and cognitive behavior therapy. Her research interests include emotion recognition, social cognition, executive functions and working memory, depression and cognitive behavior therapy. She is currently a member of APA. She has received Prof S Sinha award for Social Psychology, from University of Calcutta in 2000.